\pdfoutput=1

\documentclass[11pt]{article}

\usepackage[]{ACL2023}

\usepackage{times}
\usepackage{latexsym}
\usepackage[T1]{fontenc}
\usepackage[utf8]{inputenc}
\usepackage{microtype}

\usepackage{inconsolata}

\usepackage{amsmath,amsfonts,bm}

\def\eqref#1{equation~\ref{#1}}

\def\1{\bm{1}}

\DeclareMathAlphabet{\mathsfit}{\encodingdefault}{\sfdefault}{m}{sl}
\SetMathAlphabet{\mathsfit}{bold}{\encodingdefault}{\sfdefault}{bx}{n}

\usepackage{threeparttable}
\usepackage{multirow}
\usepackage{colortbl}
\definecolor{mygray}{gray}{0.85}

\newcommand{\ie}{i.e.}

\usepackage{xspace}
\newcommand{\vpara}[1]{\vspace{0.04in}\noindent\textbf{#1}\xspace}

\usepackage{subfigure}
\usepackage{multirow}
\usepackage[normalem]{ulem}
\usepackage[ruled,linesnumbered]{algorithm2e}

\usepackage{xspace}
\usepackage{threeparttable}
\usepackage{enumitem}
\usepackage{makecell}
\usepackage{bbding}
\usepackage{caption}
\usepackage{graphics}
\usepackage{graphicx}
\usepackage{CJKutf8}
\usepackage{subcaption} 
\usepackage{subfigure}

\usepackage[utf8]{inputenc} 
\usepackage[T1]{fontenc}    
\usepackage{url}            
\usepackage{booktabs}       
\usepackage{amsfonts}       
\usepackage{nicefrac}       
\usepackage{microtype}      
\usepackage{lipsum}
\usepackage{csquotes}
\usepackage{fancyhdr}       
\usepackage{longtable}
\usepackage{graphicx}       
\usepackage{multirow}
\usepackage{booktabs}

\title{Towards Comprehensive Preference Data Collection for Reward Modeling}

\author{\bf
Yulan Hu$^{1,2}$,
Qingyang Li$^2$,
Sheng Ouyang$^{1,2}$,
Ge Chen$^{2,3}$,
Kaihui Chen$^2$,\\
\bf
Lijun Mei$^2$,
Xucheng Ye$^2$,
Fuzheng Zhang$^2$,
Yong Liu$^1$
\\
$^1$ Renmin University of China, Gaoling School of Artificial Intelligence, Beijing \\
\texttt{huyulan,ouyangsheng,liuyonggsai@ruc.edu.cn}\\
$^2$ Kuaishou Technology, Beijing \\
\texttt{liqingyang,chenkaihui,yexucheng,zhangfuzheng@kuaishou.com}\\
$^3$ 	University of Chinese Academy of Sciences, Beijing \\
\texttt{chenge221@mails.ucas.ac.cn}\\
}

\begin{document}
\maketitle
\begin{abstract}
Reinforcement Learning from Human Feedback (RLHF) facilitates the alignment of large language models (LLMs) with human preferences, thereby enhancing the quality of responses generated. A critical component of RLHF is the reward model, which is trained on preference data and outputs a scalar reward during the inference stage. However, the collection of preference data still lacks thorough investigation. Recent studies indicate that preference data is collected either by AI or humans, where chosen and rejected instances are identified among pairwise responses. We question whether this process effectively filters out noise and ensures sufficient diversity in collected data. To address these concerns, for the first time, we propose a comprehensive framework for preference data collection, decomposing the process into four incremental steps: Prompt Generation, Response Generation, Response Filtering, and Human Labeling. This structured approach ensures the collection of high-quality preferences while reducing reliance on human labor. We conducted comprehensive experiments based on the data collected at different stages, demonstrating the effectiveness of the proposed data collection method.
\end{abstract}

\section{Introduction}
Reinforcement Learning from Human Feedback (RLHF) \cite{rlhf} has demonstrated significant potential in aligning Large Language Models (LLMs) with human preferences \cite{instructgpt, llama2}. Within the RLHF framework, the reward model (RM) outputs a scalar reward for a given prompt and response, further guiding the reinforcement learning. The reward model typically relies on collected preference data for training, enabling it to distinguish between chosen and rejected responses \cite{secret_rm}.

Recent years have seen increasing discussions on constructing and improving reward models (RMs). Notable strategies include employing mixtures of experts architectures \cite{moe} to enhance model robustness, and ensembling logits \cite{rm_ensemble_0, rm_hacking_0} or parameters \cite{rm_warm} of multiple RMs to mitigate the reward hacking problem \cite{hacking_define}. These methods refine RMs from a model perspective, while studies focusing on data aspects are largely overlooked. As revealed in \cite{secret_rm}, the preference data used for reward model training—whether off-the-shelf or collected by AI or human—often contains noise and may not be suitable for RM training. Unfortunately, both released technical reports \cite{anthropic_helpful, llama2, qwen_technique_report, baichuan2, deepseekv2} and research studies lack detailed analysis on collecting high-quality preference data for RM training.

In this paper, we present the first comprehensive study on collecting preference data for training reward models (RMs). We propose a framework designed to gather high-quality preference data. Specifically, we decompose the preference data collection process into four sub-steps: \textbf{Prompt Generation}, which selects challenging prompts that the SFT model struggles to handle; \textbf{Response Generation}, which produces diverse responses to enhance the model's generalization; \textbf{Response Filtering}, which removes noisy answer pairs; and \textbf{Human Labeling}, which annotates a modest amount of pseudo preference data. Finally, the RM is trained on the data reviewed by human labelers.

As an initial attempt to thoroughly investigate preference data collection for reward models (RMs), the proposed framework consolidates AI filtering with human intervention. Compared to relying solely on AI or human annotation, this framework effectively reflects human preferences while significantly reducing the amount of human labor required. We conducted experiments on preference data collected at different stages, demonstrating that performance enhancement is achieved as the quality of the preference data improves. This study bridges the gap in research on preference data collection for RMs.

\section{Methodology}
\begin{figure}[h]
\centering 
\includegraphics[width=0.48\textwidth]{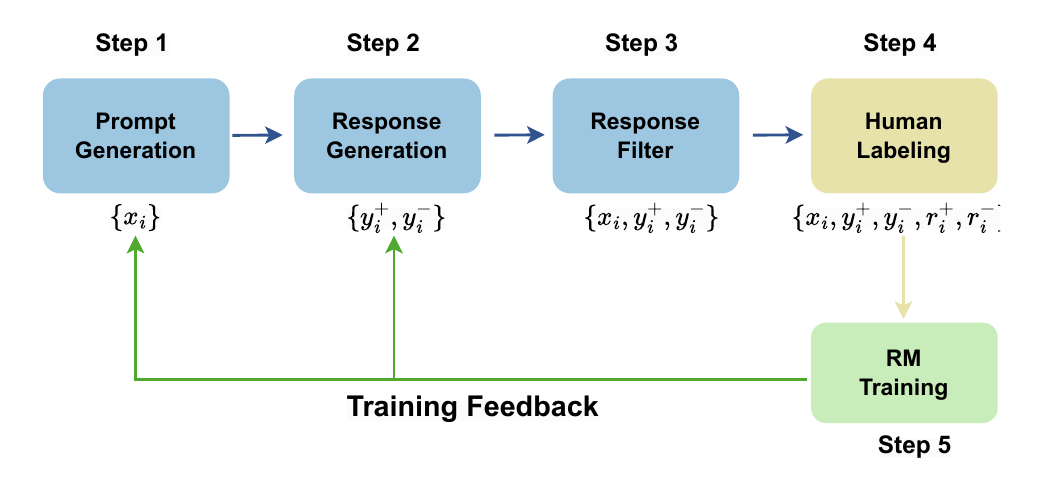}
\caption{The overview of the proposed framework.} 
\label{fig:overview}
\end{figure}
In this section, we present the details of the proposed framework. As illustrated in Figure \ref{fig:overview}, the framework comprises five hierarchical steps, with the first four steps dedicated to preference data collection. The first three steps involve intricate design, while the fourth step is primarily carried out by annotators. In the subsequent sections, we first present standard RM training process, followed by a step-by-step deconstruction of the proposed framework.

\subsection{RM Training}
RLHF \cite{instructgpt} focuses on maximizing the reward of generated samples \cite{rlhf, instructgpt}, involving a reward model (RM) that outputs a scalar reward to evaluate the quality of given text. Consider a RM \( r_\psi \) parameterized by \( \psi \), where \( r_\psi \) is initialized from a SFT model \( \pi_{SFT} \) and then trained in a supervised manner on preference data \(\mathcal{D}=\{( x_i, y_{i}^{+},  y_{i}^{-})\}_{i=1}^{N}\). Here, \( x_i \) represents the prompt, and \( y_{i}^{+} \) and \( y_{i}^{-} \) are the two responses, with \( y_{i}^{+} \) preferred over \( y_{i}^{-} \). After collecting sufficient preference data, we frame the RM training as a binary classification problem as follows:
\begin{equation}
\begin{split}
\mathcal{L}\left(r_{\psi}, \mathcal{D}\right)  =
& -\mathbb{E}_{\left(x, y^{+}, y^{-}\right) \sim \mathcal{D}} \\ & \left[\log \sigma\left(r_{\psi}\left(x, y^{+}\right) - r_{\psi}\left(x, y^{-}\right)\right)\right],
\end{split}
\label{rm:loss}
\end{equation}

\subsection{The proposed framework}
\vpara{Step 1: Prompt Generation.}
\begin{figure}[h]
\centering 
\includegraphics[width=0.5\textwidth]{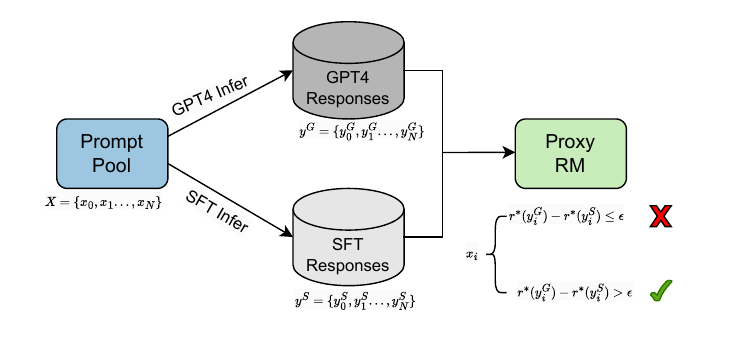}
\caption{The prompt generation (Step 1) process.} 
\label{fig:prompt_generation}
\end{figure}
The prompt generation phase aims to collect sufficient challenging prompts, which will be used to generate responses for RM training. The model trained through RLHF should preserve the overall capabilities of the SFT model while also being able to handle those prompts that are difficult for the SFT model. To achieve this, two critical issues need to be addressed in prompt generation. First, the prompt set should exhibit diversity to avoid the barrel effect, i.e., where the RM scores accurately in one domain but is less precise in another. Second, the prompt set should include samples that are difficult for the SFT model to handle.
 
To address the two issues, we develop a comprehensive strategy illustrated in Figure \ref{fig:prompt_generation}. We randomly sample a sufficient number of prompts from diverse categories to form the prompt pool \( X = \{x_0, x_1, \ldots, x_N\} \) , thereby fulfilling the first requirement. For the latter requirement, our core premise is that if the quality of the response inferred by the SFT model is close to that of strong LLM models, \ie, GPT-4, for the same prompt, it indicates that the SFT model can effectively resolve this prompt, eliminating the need for further learning in the RLHF stage, we achieve this with the assist of an off-the-shelf proxy RM, $r^{\star}$. Specifically, we first use the SFT model \( s_\phi \) and the GPT-4 model to generate two responses for each prompt in \( X \), acquiring \( y^{G} \) and \( y^{S} \). Then, we use \( r^{\star} \) to score the \( \langle \text{prompt, response} \rangle \) pairs as follows:
\begin{equation}
\Delta_{( y^{G}, y^{S} )} = \left\{
\begin{aligned}
r^{\star}(x, y^{G}) - r^{\star}(x, y^{S}) \le \epsilon, & \quad \text{drop} \, x \\
r^{\star}(x, y^{G}) - r^{\star}(x, y^{S}) > \epsilon, & \quad \text{keep} \, x \\
\end{aligned}
\right.
\label{equa:rm_score_diff}
\end{equation}
where \( \epsilon \) denotes the preset threshold difference between \( y^{G} \) and \( y^{S} \). Equation \ref{equa:rm_score_diff} indicates that we only keep the prompts if the corresponding response generated by the existing SFT model \( s_\phi \) is relatively lagging behind the well-performing models. This approach filters the "hard" prompt samples from \( X \), obtaining the refined prompt set \( X^{*} \) for RM training.

\vpara{Step 2: Response Generation.}
RM training accepts a prompt \( x \) and two preference responses \((y^{+}, y^{-})\) for pairwise learning. To generate feasible \((y^{+}, y^{-})\), similar to step 1, the quality and diversity of the responses need to be ensured. The primary challenge lies in the implicit supervisory signal inherent in \((y^{+}, y^{-})\), which means that at least \( y^{+} \) should be generated by models that are at least as superior as the current model being optimized. For instance, consider \( \pi_{SFT} \) as a 13B-size SFT model; it is necessary to use a stronger model to generate the responses, such as a 65B-size or 175B-size SFT model, GPT-4, etc.

Another key requirement lies in the diversity of the responses. We achieve this by combining results from various models. Under the premise of fulfilling the first condition, we can combine multiple LLMs with different configurations, such as different parameter settings and sizes, to generate \( (y^{+}, y^{-}) \). Additionally, off-the-shelf strong models can also be employed as a supplement.  

\vpara{Step 3: Response Filtering.}
In step 2, we generate multiple pairwise responses for each prompt, forming the training candidate set \( \mathcal{D} \), with each training instance formulated as a triad \( \langle x, y^{+}, y^{-} \rangle \). Ideally, \( y^{+} \) should exhibit a certain degree of superiority over \( y^{-} \), meaning that \( \langle x, y^{+}, y^{-} \rangle \) should not be too easy or too hard for pairwise learning. However, such a condition cannot always be fulfilled. For instance, the responses to an objective question may be identical, offering no supervisory signal for RM training while conversely introducing additional labeling overhead for annotators.

Thus, refinement is necessary before sending $\mathcal{D}$ to the annotators. We incorporate GPT-4 to help filter out useless training samples. Specifically, we score each instance in $\mathcal{D}$ using the in-context learning technique \cite{icl_survey}. We divide the scoring criteria for each instance into five levels, where 1 represents the worst response quality and 5 represents the best, we then employ GPT-4 to score $x, y^{+}$ and $x, y^{-}$, respectively. It is worth noting that since the prompts belong to different categories, the corresponding scoring criteria for prompts of different categories are also different. 

After scoring, we obtain score pairs as $\langle x, y^{+}, r^{+} \rangle$ and $\langle x, y^{-}, r^{-} \rangle$. Based on these scores, we select the pairs that exhibit certain differences for the annotators to further review. We build a filtering strategy presented in Table \ref{tab:response_filter}, where we consider two kinds of samples that should be discarded. First, responses with identical scores, as such pairs cannot provide any discriminative knowledge during RM training. Second, responses that exhibit extreme distinctness, for example, pairs where $r^{+}$ is scored 5 and $r^{-}$ is scored 1. We consider that the RM possesses the ability to distinguish samples with significant divergence, thus eliminating the need for further assessment by the annotators. These two kinds of hollow samples occupy a large portion of $\mathcal{D}$, and filtering them enhances labeling efficiency to a large extent.
 

\begin{table}[]
\centering
\begin{tabular}{@{}c|ccccc@{}}
\toprule
$<r^{+}, r^{-}>$ & 1                          & 2                           & 3                          & 4                          & 5                          \\ \midrule
1                               & {\color[HTML]{C0C0C0} 1-1} & {\color[HTML]{009901} 1-2} & {\color[HTML]{009901} 1-3} & {\color[HTML]{C0C0C0} 1-4} & {\color[HTML]{C0C0C0} 1-5} \\
2                               & {\color[HTML]{009901} 2-1} & {\color[HTML]{C0C0C0} 2-2}  & {\color[HTML]{009901} 2-3} & {\color[HTML]{009901} 2-4} & {\color[HTML]{C0C0C0} 2-5} \\
3                               & {\color[HTML]{009901} 3-1} & {\color[HTML]{009901} 3-2}  & {\color[HTML]{C0C0C0} 3-3} & {\color[HTML]{009901} 3-4} & {\color[HTML]{009901} 3-5} \\
4                               & {\color[HTML]{C0C0C0} 4-1} & {\color[HTML]{009901} 4-2}  & {\color[HTML]{009901} 4-3} & {\color[HTML]{C0C0C0} 4-4} & {\color[HTML]{009901} 4-5} \\
5                               & {\color[HTML]{C0C0C0} 5-1} & {\color[HTML]{C0C0C0} 5-2}  & {\color[HTML]{009901} 5-3} & {\color[HTML]{009901} 5-4} & {\color[HTML]{C0C0C0} 5-5} \\ \bottomrule
\end{tabular}
\caption{The responses filtering scoring matrix is formulated in the shape of $5 \times 5$. The scoring pairs highlighted in {\color[HTML]{009901} green} are preserved, while those in {\color[HTML]{C0C0C0} grey} are discarded.}
\label{tab:response_filter}
\end{table}

\subsection{Data Funnel}
In step 4, the annotators further review and score the filtered training samples from step 3. The ultimately reviewed results will be used for RM training in step 5.

\begin{figure}[h]
\centering 
\includegraphics[width=0.48\textwidth]{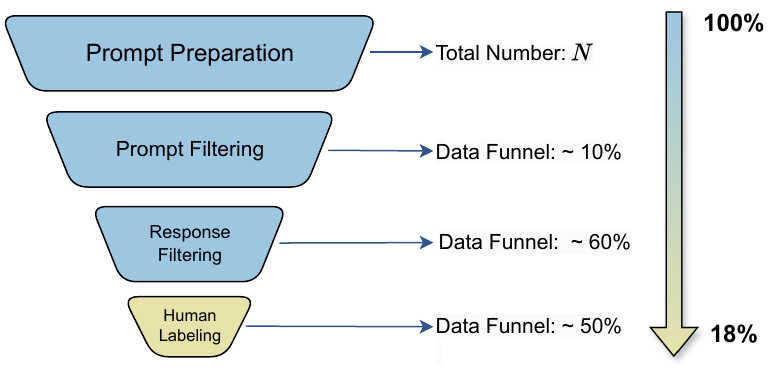}
\caption{The data funnel illustrates the loss rate at each step. The ultimate data retention is roughly 20\%, meaning that only 20\% of the prepared samples are qualified for training.}
\label{fig:data_funnel}
\end{figure}

As a hierarchical RM data generation method, the proposed framework involves multiple filtering strategies. Consequently, a data funnel exists between each pair of steps, meaning that not all data from the previous step will be fully transferred to the next step. We illustrate the practical data funnel in Figure \ref{fig:data_funnel}. Consider that the initial number of candidate prompts is $N$. The loss rate of the prompt filter (Step 1) is relatively small, around 10\%, while the loss rate for Step 3 is comparatively large, nearly 60\%. Finally, after human labeling, we discard nearly half of the labeled samples from step 3 that are not appropriate for training. Overall, around 20\% of the prepared training samples are filtered for RM training.

\section{Experiment}
\subsection{Setups}
We employ two SFT models of varying sizes (13B and 65B) as the base model, the basic architecture are built upon LLaMA~\cite{llama}. The overall preference data are collected from two sources: the available open-source preference data and the data inferred by AI or human, each accounts for about 30w. To reveal the effectiveness and necessity of different steps of the proposed framework, we follow the prompt preparation step to collect roughly 15w refined prompts, then proceed with the proposed preference data collection procedure. Specifically, we train two RMs based on the data collected at step 3 and step 4, respectively. Roughly, around 5.4w preference data is collected in step 3 and around 3w preference data are finally used in step 4.

We save the checkpoint of the last iteration for evaluation. We evaluate the two RMs from two aspects, the preference benchmarks and the overall performance of the RMs incorporation with downstream policy. We follow the preference benchmark used in~\cite{llama2, qwen_technique_report}, containing Anthropic Helpfulness~\cite{anthropic_helpful}, OpenAI Summarize~\cite{openai_summ}, OpenAI WebGPT~\cite{openai_webgpt} and Standford SHP~\cite{stanford_shp}.
 
\subsection{Results}
\vpara{Results on preference benchmarks.}
We report the results on preference benchmarks in Table \ref{tab:main_results}, using accuracy as the evaluation metric. The results for both the 13B-size and 65B-size RMs validate the improvement from step 3 to step 4, indicating that refinement of preference data can indeed boost performance. Despite the scale of preference data used in step 3 being almost twice that used in step 4, we observe that the refinement of data quality is beneficial.
\begin{table}[]
\resizebox{\linewidth}{!}{
\begin{tabular}{@{}ccccccc@{}}
\toprule
\multicolumn{2}{c}{RMs}         & \begin{tabular}[c]{@{}c@{}}Anthropic\\ Helpful\end{tabular} & \begin{tabular}[c]{@{}c@{}}OpenAI\\ Summ.\end{tabular} & \begin{tabular}[c]{@{}c@{}}Stanford\\ SHP\end{tabular} & WebGPT & Overall \\ \midrule
\multirow{2}{*}{13B} & RM-Step3 & 68.7                                                        & 68.2                                                   & 67.2                                                   & 65.9   & 67.5    \\
                     & RM-Step4 & 69.6                                                        & 68.6                                                   & 68.1                                                   & 66.7   & 68.3    \\ \midrule
\multirow{2}{*}{65B} & RM-Step3 & 69.9                                                        & 68.7                                                   & 71.5                                                   & 71.4   & 70.4    \\
                     & RM-Step4 & 71.4                                                        & 71.4                                                   & 72.1                                                   & 70.8   & 71.4    \\ \bottomrule
\end{tabular}}
\caption{The results on preference benchmarks.}
\label{tab:main_results}
\end{table}

\begin{figure}[htp]
    \centering    
        \subfigure[13B]{\includegraphics[width=0.235\textwidth]{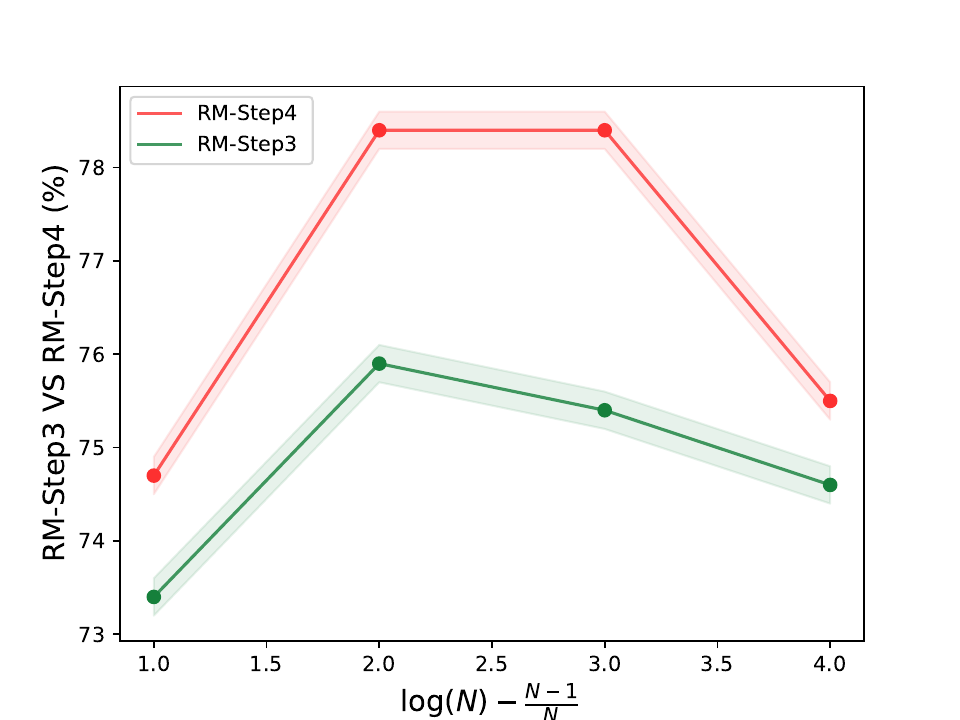}}
        \subfigure[65B]{\includegraphics[width=0.235\textwidth]{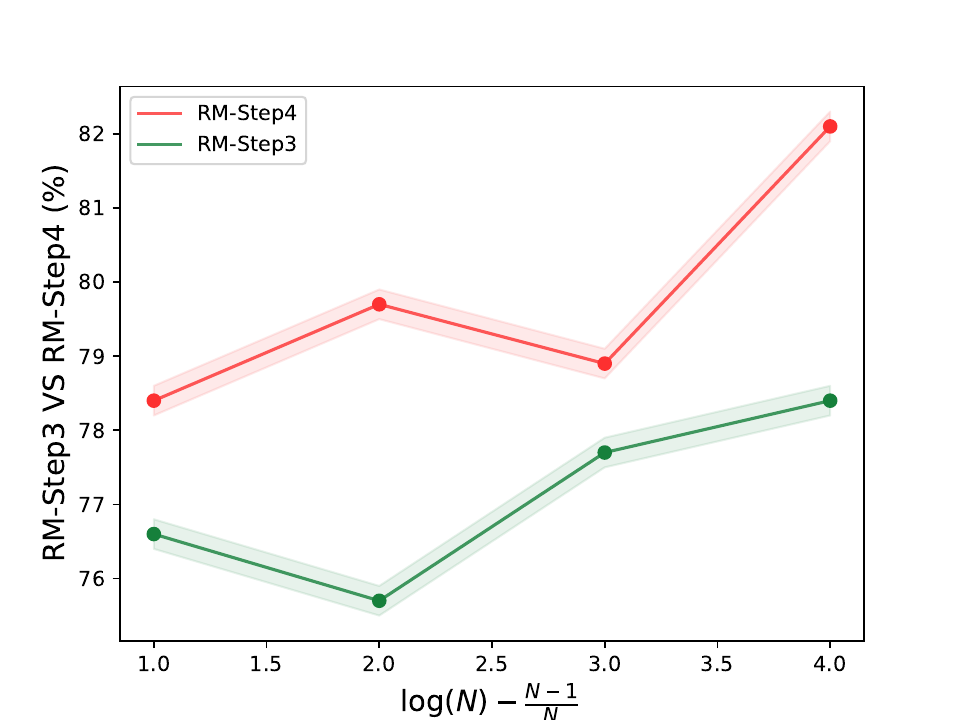}}
    \caption{Win rates of the reward model trained with preference data in Step 4 against Step 3.}
    \label{fig:bon_res}
\end{figure}

\vpara{Results of Best-of-N experiments.} In addition, we integrate the trained RMs with the BoN reranking policy~\cite{raft}. BoN is an inference-time sampling strategy that aims to select the answer with the highest reward from $n$ candidates, usually generated by the SFT model $\pi_{SFT}$. The gains obtained by BoN are approximated by $\text{log}(N) - \frac{N-1}{N}$~\cite{bon_theoretical}. Our BoN experiments are conducted on AlignBench~\cite{alignbench}. For each prompt in AlignBench, we use the SFT model to generate $n$ answers and choose the best answer from the answer set based on the RM score. The value of $n$ is chosen from \{5, 10, 20, 50\}. We then calculate the win rate for the RM trained on preference data collected in step 3 against step 4, and plot the results in Figure~\ref{fig:bon_res}. The results validate that the reward models consistently help select better answers than the raw SFT model for both the 13B and 65B models, further verifying the enhancement in performance with the refinement of preference data.

\section{Conclusion}
In this paper, we conduct a thorough inspection and develop a framework for the collection of preference data for RM training. Specifically, we decompose the process into several sub-steps, which facilitates the collection of high-quality data while reducing the labor required from humans. We validate the framework using both preference data benchmarks and policy learning, with results demonstrating improvements in data quality brought about by the framework. As an initial attempt, we believe the proposed framework bridges the gap in comprehensive preference data collection within the LLM community.

\newpage
\section{Limitations}
We discuss the limitation of the proposed framework in this section, namely, the relatively long-term preference data production pipeline.

\vpara{Long-term data production.} As illustrated in Figure~\ref{fig:overview}, the proposed framework contains four steps to obtain the ultimate high-quality preference data, each step requires relatively extensive filtering. The long-term collection pipeline may not facilitate collect enough training data in a short period of time. Therefore, we believe the proposed framework is more suitable for the later stages of RM optimization and for optimizing certain specific verticals. In the early stages, we can rely on AI or open-source data for RM tuning.

\bibliography{ref}
\bibliographystyle{acl_natbib}

\end{document}